\documentclass[10pt,twocolumn,letterpaper]{article}

\usepackage{cvpr}              

\usepackage{algorithm}
\usepackage{algorithmic}
\usepackage{amsmath}
\usepackage{bbding}
\usepackage{booktabs}
\newcommand\norm[1]{\left\lVert#1\right\rVert}
\newcommand\round[1]{\left[#1\right]}
\usepackage{multirow}
\usepackage{arydshln}
\usepackage[accsupp]{axessibility}
\usepackage{xcolor}
\usepackage{amsfonts}
\usepackage{graphicx}
\usepackage[pagebackref,breaklinks,colorlinks]{hyperref}

\usepackage[capitalize]{cleveref}
\crefname{section}{Sec.}{Secs.}
\Crefname{section}{Section}{Sections}
\Crefname{table}{Table}{Tables}
\crefname{table}{Tab.}{Tabs.}

\def\cvprPaperID{4560} 
\def\confName{CVPR}
\def\confYear{2022}

\begin{document}

\title{Motion-aware Contrastive Video Representation Learning via \\ Foreground-background Merging}
\author {
    Shuangrui Ding\textsuperscript{\rm 1}\thanks{Work done during an internship at Tencent AI Lab.} \quad
    Maomao Li\textsuperscript{\rm 2} \quad Tianyu Yang\textsuperscript{\rm 2}  \quad Rui Qian\textsuperscript{\rm 3} \\Haohang Xu\textsuperscript{\rm 1}  \quad Qingyi Chen\textsuperscript{\rm 4} \quad Jue Wang\textsuperscript{\rm 2} \quad 
    Hongkai Xiong\textsuperscript{\rm 1}\thanks{ Corresponding author. Email: xionghongkai@sjtu.edu.cn.}\\
    \textsuperscript{\rm 1}Shanghai Jiao Tong University \quad 
    \textsuperscript{\rm 2}Tencent AI Lab  \\  \textsuperscript{\rm 3}The Chinese University of Hong Kong  \textsuperscript{\rm 4}University of Michigan \\
    {\tt\small \{dsr1212, xuhaohang, xionghongkai\}@sjtu.edu.cn tianyu-yang@outlook.com}\\
    {\tt\small \{limaomao07, arphid\}@gmail.com  qr021@ie.cuhk.edu.hk chenqy@umich.edu}
}

\maketitle


\begin{abstract}
In light of the success of contrastive learning in the image domain, current self-supervised video representation learning methods usually employ contrastive loss to facilitate video representation learning. When naively pulling two augmented views of a video closer, the model however tends to learn the common static background as a shortcut but fails to capture the motion information, a phenomenon dubbed as background bias. Such bias makes the model suffer from weak generalization ability, leading to worse performance on downstream tasks such as action recognition. To alleviate such bias, we propose \textbf{F}oreground-b\textbf{a}ckground \textbf{Me}rging (FAME) to deliberately compose the moving foreground region of the selected video onto the static background of others. Specifically, without any off-the-shelf detector, we extract the moving foreground out of background regions via the frame difference and color statistics, and shuffle the background regions among the videos. By leveraging the semantic consistency between the original clips and the fused ones, the model focuses more on the motion patterns and is debiased from the background shortcut. Extensive experiments demonstrate that FAME can effectively resist background cheating and thus achieve the state-of-the-art performance on downstream tasks across UCF101, HMDB51, and Diving48 datasets. The code and configurations are released at \href{https://github.com/Mark12Ding/FAME}{https://github.com/Mark12Ding/FAME}.
\end{abstract}

\section{Introduction}
\label{intro}
The recent development of deep learning has promoted a series of applications in videos, such as video recognition~\cite{wang2016temporal, tran2018closer, feichtenhofer2019slowfast}, video retrieval~\cite{zhang2018cross, gabeur2020multi}, and video object segmentation~\cite{Croitoru_2017_ICCV, Zhang_2020_CVPR, Huang_2020_CVPR}.
While various large-scale benchmarks~\cite{abu2016youtube,carreira2017quo, goyal2017something} are the key to those successes, the costly manual annotation involved in fully-supervised methods excludes the potential utilization of millions of uncurated videos on the Internet.
To further advance the video-related research, learning video representation in an unsupervised manner is of great significance and emerges as a general trend in the computer vision community.

\begin{figure}
    \centering
    \subfloat[Results of vanilla contrastive learning.]{%
      \includegraphics[width=0.1\textwidth]{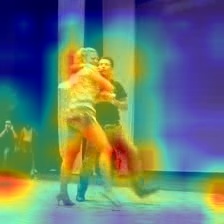}
      \includegraphics[width=0.1\textwidth]{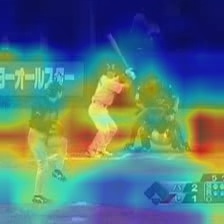}
      \includegraphics[width=0.1\textwidth]{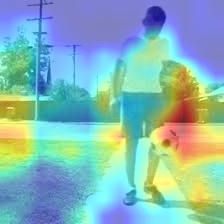}
      \includegraphics[width=0.1\textwidth]{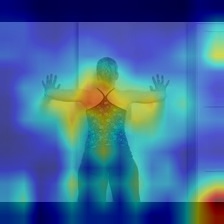}
    }\\\vspace{0.2em}
    \subfloat[Results of our approach FAME.]{%
      \includegraphics[width=0.1\textwidth]{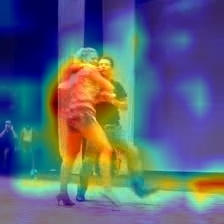}
      \includegraphics[width=0.1\textwidth]{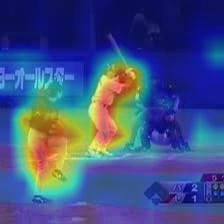}
      \includegraphics[width=0.1\textwidth]{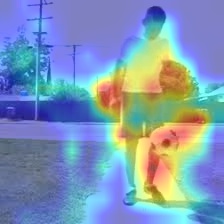}
      \includegraphics[width=0.1\textwidth]{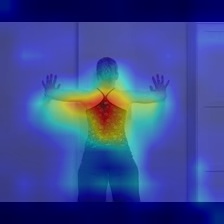}
    } 
    \caption{Class-agnostic activation map~\cite{baek2020psynet} visualization of important areas. The heatmap indicates how much the pretrained model attends to the region. Compared to the conventional approach, our method mitigates the background bias significantly.}
    \label{fig:caam}
    \vspace{-0.5em}
\end{figure}

\begin{figure}
    \centering
    \includegraphics[width=0.4\textwidth]{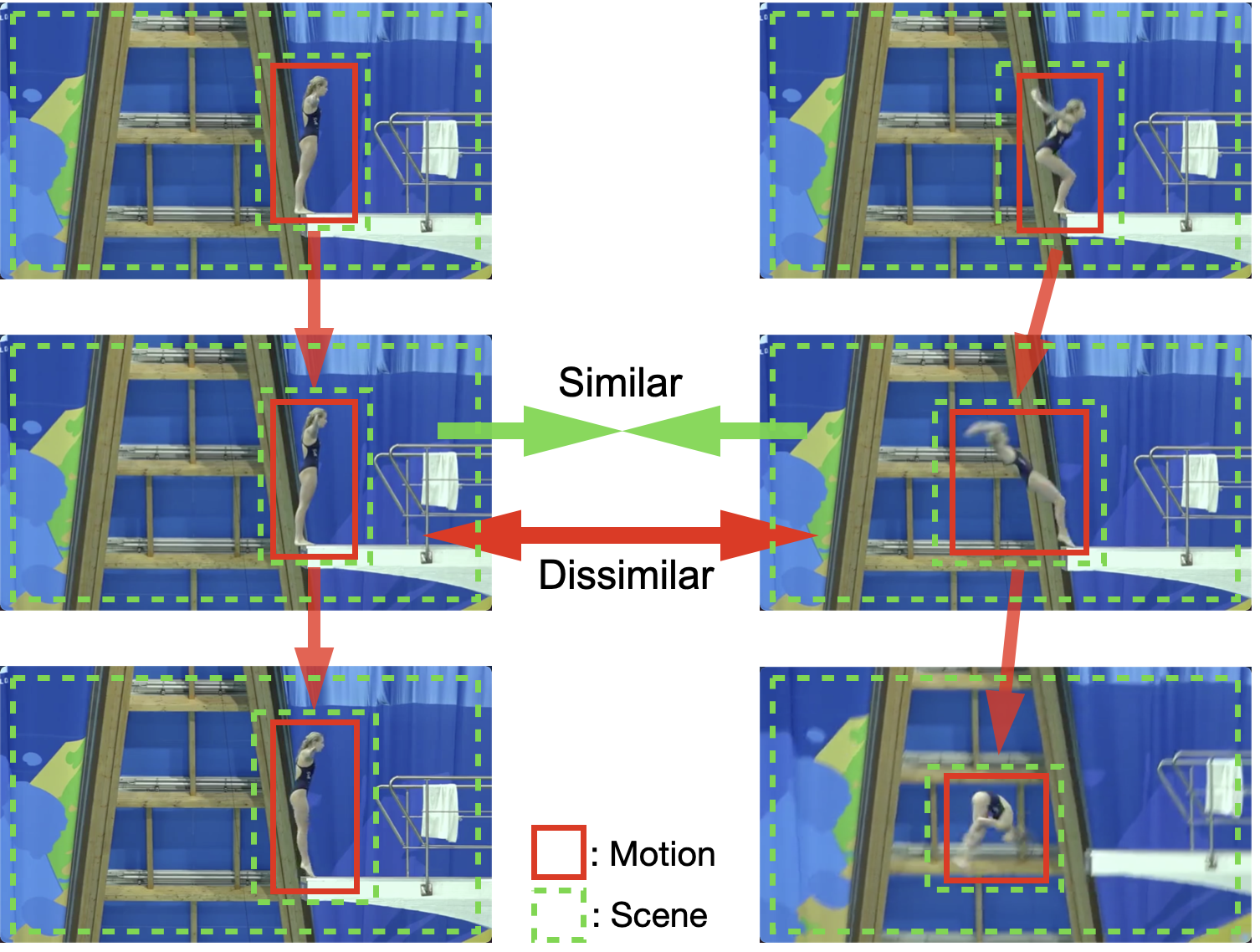}
    \caption{Illustration on a diving sequence. The green dashed box represents the scene and the red box means motion area. The two clips have the same background but distinct motions. Drawing such positive pairs closer inclines the model towards static bias.}
    \label{fig:teaser}
    \vspace{-0.5em}
\end{figure}

Recently, unsupervised learning in images~\cite{oord2018representation,wu2018unsupervised,tian2019contrastive,chen2020exploring} has achieved competitive performances compared to their supervised counterparts, especially with the contrastive self-supervised learning formulation~\cite{he2020momentum, chen2020simple}. The common idea of contrastive learning is to pull `positive' pairs together in the embedding space and push apart the anchor from `negative' samples. Due to inaccessibility to the label, a positive pair is usually formed by data augmentations of the anchor sample while the negative samples come from other samples. Inspired by these successes, various attempts have also been made in self-supervised video representation learning~\cite{qian2021spatiotemporal, feichtenhofer2021large}. 
However, we find applying vanilla contrastive learning on the video domain directly will lead the model to attending to the static area. As illustrated in Fig.~\ref{fig:caam}, vanilla contrastive learning does not concentrate on the moving actors or objects but lays much emphasis on background areas. There might be two possible reasons: 1) the background usually covers much more area than the moving objects in the whole video, so that the model is more likely to focus on the background. 2) when sampling two different clips of the video, the static contexts are almost the same but there tend to be subtle differences in motion patterns. We show an example in Fig.~\ref{fig:teaser}. There are two clips sampled from one diving video. The green region is the background, which occupies over 3/4 area. And the red box, a small area, contains the moving diver. In addition, 
the backgrounds of the two clips are almost identical while two motions appear somewhat different, i.e., one is standing on the springboard, the other is taking off.
That is to say, when we follow conventional visual augmentation techniques to form the positive pair and employ the multi-view constraint as self-supervision, it is intuitive for the model to pull static features close but attend less to motions. Hence, to let the contrastive learning pipeline be more motion-aware, we need to construct positive pairs in the way where the motions are more similar than backgrounds. 

Here comes to a question on how to fabricate the motion-aware positive pair that makes contrastive learning prioritize the motion patterns. This paper explores this feasibility and presents a new augmentation technique named \textbf{F}oreground-b\textbf{a}ckground \textbf{Me}rging (FAME). Our motivation is to keep motion areas (foreground) unchanged as much as possible and replace static areas (background) with irrelevant content. 
Specifically, we first circle out the edge region of the moving object as the seed region via frame difference. Then, we use color statistics to extrapolate the entire moving object from the seed region. This efficient foreground discovery method extracts dynamic areas on which we expect the model to put emphasis. Then, we fuse the extracted foreground regions of each video with random backgrounds from other videos to form new action samples. In this way, when we force the model to learn the consistent representation between original clips and distracting clips, the model has to learn representations that are sensitive to motion patterns and overcome the background cheating. 
We evaluate the proposed FAME on three action recognition benchmarks. The superior experimental performance verifies that FAME enables self-supervised contrastive video representation learning to generalize better and distill the motion-aware representations. 
In short, we summarize our contributions as follows:
\begin{itemize}
    \item We demonstrate the background bias caused by vanilla contrastive learning and propose a simple yet effective augmentation method FAME to help the model break background shortcuts and learn motion-aware representations. 
    \item Our method enhances the conventional contrastive learning without whistle and bell and achieves the start-of-the-art performance among UCF101, HMDB51, and Diving48 datasets. 
\end{itemize}

\section{Related Work}
\noindent{\textbf{Contrastive Visual Representation Learning.}}
Recently, contrastive learning has greatly facilitated self-supervised visual representation learning~\cite{wu2018unsupervised,oord2018representation,tian2019contrastive,chen2020simple,he2020momentum}. It performs instance discrimination in a fully self-supervised manner to pull the representations of the same instance close and push those of different instances far away. Following this idea, \cite{wu2018unsupervised} proposes to formulate the instance discrimination as a non-parametric classification problem.  
\cite{oord2018representation} mathematically proves that we could estimate mutual information with InfoNCE loss~\cite{gutmann2010noise}, which can be easily used for optimization. Later, MoCo~\cite{he2020momentum} proposes to make use of key representations calculated in previous iterations as negative samples to facilitate contrastive learning. SimCLR~\cite{chen2020simple} employs a large batch size instead of the memory bank to expand the negative pool for more robust visual representation. Considering that SimCLR requires tremendous computational resources, we adopt the MoCo framework as a strong baseline for self-supervised pretraining in our work. 

\noindent{\textbf{Self-supervised Video Representation Learning.}}
In video representation learning, there has been a line of works that employ diverse pretext tasks for self-supervised representation learning~\cite{misra2016shuffle,lee2017unsupervised,xu2019self}. The most prevalent approaches include temporal order prediction~\cite{misra2016shuffle, xu2019self}, video colorization~\cite{vondrick2018tracking}, spatio-temporal puzzling~\cite{kim2019self} and speed prediction~\cite{benaim2020speednet}. These methods generally employ manually designed tasks to seek the spatio-temporal cues in video data, but the performance is limited. Then, for further improvement, some works apply contrastive learning formulation into video representation learning~\cite{feichtenhofer2021large,qian2021spatiotemporal}. Han et al. use InfoNCE loss to guide dense predictive coding in videos~\cite{han2019video,han2020memory}. Based on the contrastive formulation, \cite{huang2021ascnet,chen2021rspnet} jointly learn appearance and speed of videos, and~\cite{zhang2021incomplete} simultaneously encodes inter- as well as intra-variance in videos. \cite{han2020self,recasens2021broaden, alwassel2019self} propose to leverage the consistency between different modalities to enhance video representation.
Our method focuses only on the single modality, i.e., raw RGB video, to explicitly construct positive samples with the same motions but different backgrounds for self-supervised contrastive video representation learning.

\noindent{\textbf{Video Background Bias Mitigation.}}
How to mitigate the background bias has been a long-standing topic~\cite{he2016human, li2018resound, choi2019sdn, wang2018pulling} for action recognition. 
In the supervised scenario, \cite{choi2019sdn} uses an off-the-shelf human detector to mask out the human regions and train the model in an adversarial manner. \cite{li2018resound} proposes a procedure to reassemble existing datasets that alleviates static representation bias. Later, to make the self-supervised video representations more robust to the background bias, a line of works employ other natural supervision~\cite{huang2021self, xiao2021modist, Li_2021_ICCV} to guide the model to capture motion information explicitly. However, these methods require more than one backbone to pretrain multi-modality data, resulting in an undesired computational cost. To better utilize the implicit motion information in videos, DSM~\cite{wang2021enhancing} aims to decouple the motion and context by deliberately constructing the positive/negative samples through spatial and temporal disturbance. BE~\cite{wang2021removing} proposes to add a static frame as background noise for static bias mitigation. However, these two methods would erode the moving objects and impair the motion patterns. In contrast, our method solves this shortcoming by meticulously extracting dynamic foreground regions and preserving high-quality motion patterns.

\noindent{\textbf{Copy-paste Augmentation.}}
Copy-paste augmentation~\cite{dwibedi2017cut, dvornik2018modeling, ghiasi2021simple} is a simple way to combine information from various instances and has been proved to be a good match for object-aware learning.  In addition, Mixup~\cite{zhang2018mixup} and CutMix~\cite{yun2019cutmix} share a similar idea to increase the robustness against input corruptions.
Inspired by these successes in supervised learning, MixCo~\cite{kim2020mixco} applied Mixup into visual contrastive learning and construct the semi-positive image from the mix-up of positive and negative images. Besides, InsLoc~\cite{yang2021instance} proposes to copy image instances and paste them onto background images at diverse locations and scales, which advances self-supervised pretraining for object detection. 
FAME also copy and paste foreground content onto another video, a bit like CutMix.  
But one key difference in our work compared to CutMix~\cite{yun2019cutmix} is that we leverage motion inductive bias to guide the extraction of the foreground region. Therefore, we can guarantee synthesized sample contains motion information rather than a random patch like CutMix.

\begin{figure}
    \centering
    \includegraphics[width=0.45\textwidth]{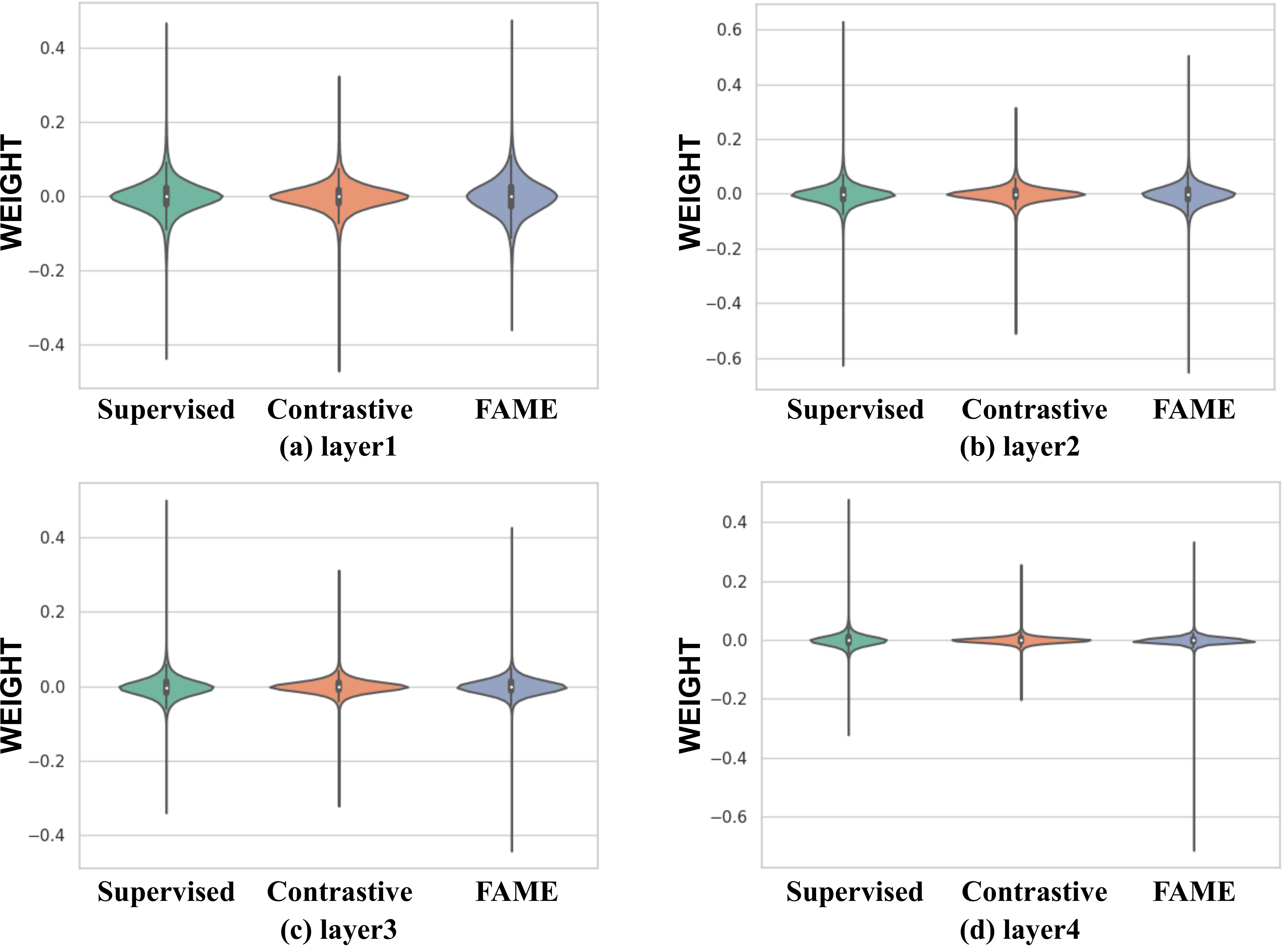}
    \caption{The statistics of temporal kernel weights at all layers of R(2+1)D. The learned kernel weights in the supervised/contrastive/FAME manner are violin-plotted from left to right.}
    \label{fig:violin}
    \vspace{-0.5em}
\end{figure}

\section{Approach}
In this section, we introduce our Foreground-background Merging (FAME) method. In section 3.1, we first revisit the vanilla contrastive learning framework based on instance discrimination~\cite{wu2018unsupervised} 
and shed light on the background bias when vanilla contrastive learning is transferred into the video domain. In section 3.2, we elaborate on how to separate foreground regions using our method. 
To clarify the notation, we denote the video clips as $X \in R^{C\times T \times H  \times W}$, where $C, T, H, W$ represents the dimension of the channel, timespan, height, width, respectively.

\subsection{Background Bias in Contrastive Learning}

The vanilla contrastive learning approach employs instance discrimination to learn the feature representation in a fully self-supervised manner~\cite{chen2020simple,he2020momentum,grill2020bootstrap}. Generally, it aims to maximize the similarity between the query sample $q$ and its positive keys $k^+$, and minimize the similarity between $q$ and negative keys $k^-$. We empirically use InfoNCE loss~\cite{gutmann2010noise} for optimization:
\begin{align}
    \mathcal{L}_{nce} = -\log\frac{\sum_{k\in\{k^+\}}\exp(\text{sim}(q,k)/\tau)}{\sum_{k\in\{k^+,k^-\}}\exp(\text{sim}(q,k)/\tau)},
    \label{eq:infonce}
\end{align}
where $\tau$ is the temperature hyper-parameter controlling the concentration level of the distribution, and $\text{sim}(q,k)$ measures the cosine similarity between the latent embeddings, i.e., $\text{sim}(q,k)=q^Tk/(\norm{q}_2\norm{k}_2)$. In most existing works~\cite{feichtenhofer2021large}, $k^+$ is the set of clip embeddings extracted from the same video as $q$, and $k^-$ is the set from other videos. 

\begin{figure}[t]
    \centering
    \includegraphics[width=0.85\linewidth]{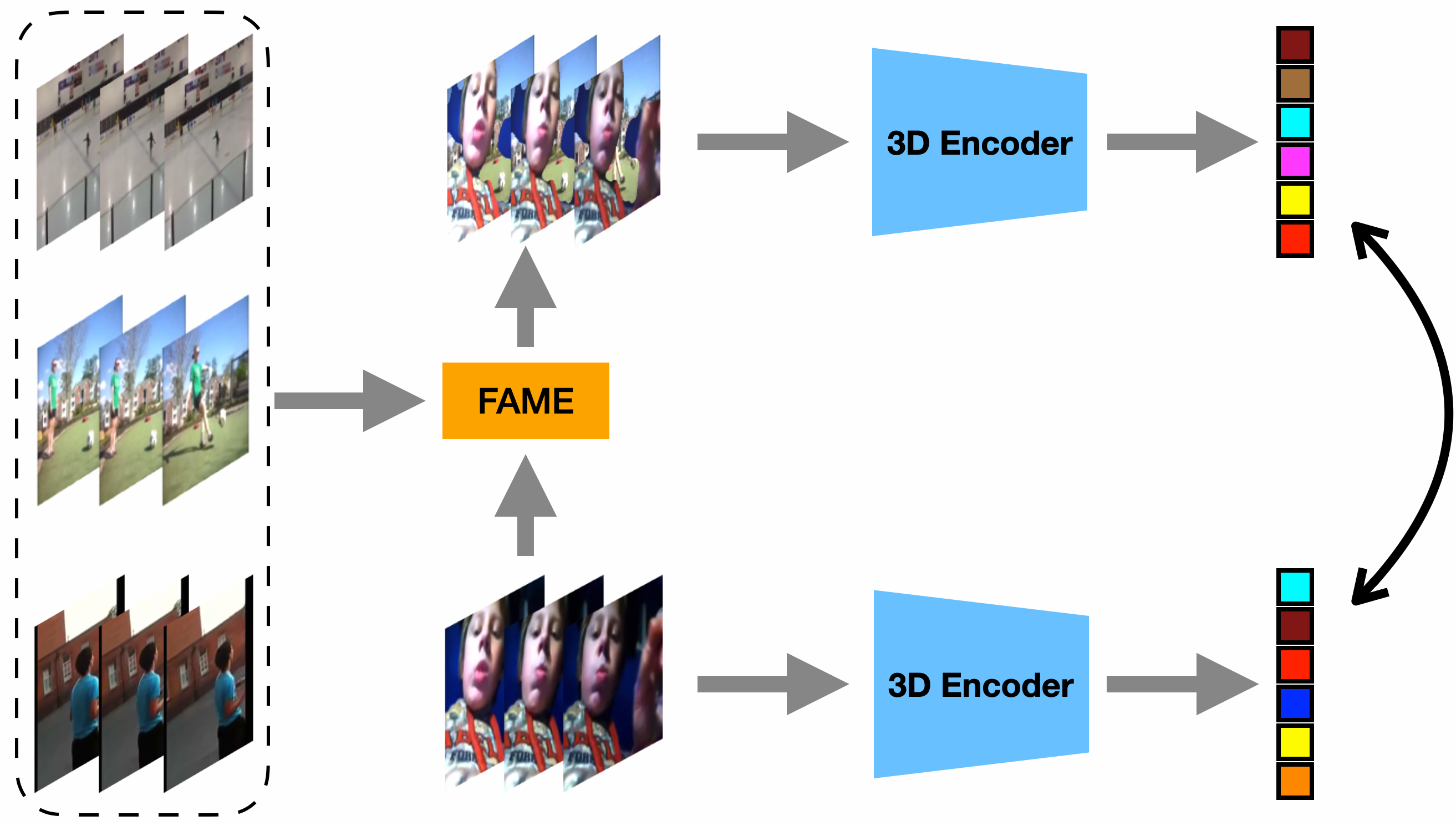}
    \caption{The contrastive learning framework with the proposed FAME. We first randomly sample two clips from a video and use FAME to generate new clips by composing the original foreground onto various backgrounds from other videos. Then, we feed the augmented clips into the existing contrastive learning scheme and perform self-supervised pretraining.}
    \label{fig:moco}
    \vspace{-0.5em}
\end{figure}

However, this vanilla contrastive learning formulation in the video domain cannot fully utilize the dynamic motion information and tends to discriminate different instances according to the background cues~\cite{wang2021removing}. To demonstrate this phenomenon, we plot the 1D convolution layers' kernel weights of R(2+1)D~\cite{tran2018closer} trained by both supervised manner~\footnote{The supervised pretrained R(2+1)D is from torchvision library.} and contrastive manner. As depicted in Fig.~\ref{fig:violin}, the weights learned via contrastive formulation are more compact and clustered at all layers than the weights learned under supervision. 
It reveals that the supervised model allows more flexible temporal modeling, while the contrastive-based counterpart presents less temporal diversity and prefers static cues to temporal dynamics.
Moreover, to consolidate our findings, we adopt the class-agnostic activation map (CAAM)~\cite{baek2020psynet} to measure the spatial attention in that CAAM can fairly evaluate pretrained representations without additional training. As shown in Fig.~\ref{fig:caam}(a), the model trained by the traditional contrastive task cannot capture the moving object correctly and is distracted by the static background. This phenomenon further indicates that there exists the static background bias in the positive pair formulation. As mentioned in Sec.~\ref{intro}, two temporally different clips usually own similar static backgrounds but distinct motion patterns. Thus, when simply pulling two augmented clips closer, the model leans to prioritize the background alignment and give up grasping the dynamic motion. 
To deal with it, we carefully design FAME as an augmentation technique. Our idea is simple. We erase the static areas on purpose and retain the dynamic areas to construct the positive pairs. By doing so, the model has to align the motion area firstly and break the static shortcut. We show the contrastive learning framework with the proposed FAME in Fig.~\ref{fig:moco}. In detail, we randomly sample two clips from different timestamps. Before applying the basic augmentation, we use our proposed FAME method to compound the foreground of one clip with the background from other videos in the same mini-batch. After that, the two clips are more similar in motions than backgrounds. Then, we feed these two clips into the 3D encoder and treat them as the positive keys while the rest of the clips serve as negative keys. Finally, we minimize the InfoNCE loss to pretrain the 3D encoder. By constructing the positive pair with the same foreground but diverse backgrounds, we guide the model to focus on temporal cues and suppress the impact of the background.


\subsection{Foreground-background Merging}

Motivated by mitigating background bias in self-supervised video representation learning, we intend to retain the foreground regions in original videos and shuffle the background areas among various videos. To achieve this goal, we propose the Foreground-background Merging method to augment the clips with minimal computation overhead. 
Concretely, FAME firstly separates the dynamic region of the static area and then composes the foreground on the other backgrounds. 

We first differentiate adjacent frames iteratively and then sum up the magnitude of the difference along channel and timespan dimensions to generate the seed region $S$. We formulate $S \in \mathbb{R}^{H  \times W}$ as
\begin{equation}
    S = \frac{1}{T-1}\sum_{c=1}^{C}\sum_{t=1}^{T-1} \norm{X_{c,t+1}-X_{c,t}}_1.
    \label{seed}
\end{equation}
Intuitively, frame difference delivers natural dynamic motion that moving foreground objects tend to possess a great magnitude, while the static backgrounds are minor in this metric. Apart from frame difference, we also consider other methods to convey motion information like optical flow~\cite{zach2007duality}. But we find the extraction of the dense optical flow of each frame is time-consuming and the frame difference can be an ideal substitute for reducing the computational cost. In practice, we find that the large values of the seed region $S$ usually correspond to the moving objects' edge region.  
To expand the edge of the foreground objects, we take inspiration from the unsupervised foreground discovery~\cite{stretcu2015multiple} for seed propagation. Specifically, we leverage the color
distributions to estimate the entire object. Denoting $N^{(F)}$ as the total number of pixels in the foreground region and $N_{x}^{(F)}$ as the number of the given color $x$ appearing in the foreground region, the probability of a given color $x$ appearing in the foreground region can be estimated as $P(x\mid F)=N_{x}^{(F)}/N^{(F)}$. 
Similarly, the probability of $x$ belonging to the background region is $P(x\mid B)=N_{x}^{(B)}/N^{(B)}$. 
In practice, we sample the foreground color distribution in the top $50\%$ of seed region $S$ and the background color distribution in the last $10\%$ of seed region $S$. Namely, in our setting, $N^{(F)}=\round{0.5\times H \times W}$ and $N^{(B)}=\round{0.1 \times H \times W}$. Given the above two distributions for the color $x$ and the assumption that all pixels with the same color have the same probability of being the foreground and background, we 
approximate the foreground likelihood for a given color $x$ as $\displaystyle P(F \mid x)= P(x \mid F)/[P(x \mid F)+P(x \mid B)].$ Therefore, the soft segmentation mask $\displaystyle M \in \mathbb{R}^{H  \times W}$ can be calculated based on the color of each pixel. We formulate it as $\round{M}_{ij} = \displaystyle P(F \mid x_{ij})$, where $x_{ij}$ is the color at pixel $(i,j)$. To better filter out the background region, we binarize the mask as follows:
\begin{equation}
    \displaystyle \round{\widetilde{M}}_{ij} = 
    \left\{
    \begin{aligned}
    1 & \text{,  if} \round{M}_{ij} \text{is among Top-$\round{\beta H W}$ of $M$,}   \\
    0 & \text{, otherwise,}
    \end{aligned}
    \right.
\end{equation}
where $\beta \in [0,1]$ is a hyper-parameter to describe the portion of the foreground. For the sake of computational efficiency, the mask we generate is constant with respect of timespan $T$. We view video clips as ``image'' when counting the color statistics, i.e., $\widetilde{X}=\sum_{t=1}^{T} X_{t}/T$. 
Having foreground mask $\widetilde{M}$, we then fill the rest with a random background. Denoting $X,Y$ as foreground and background source clips, the synthetic clip 
 \begin{equation}
     X_{\text{merge}} = X\otimes\widetilde{M}+Y\otimes(1-\widetilde{M}),
 \end{equation}
where $\otimes$ is the element-wise multiplication. Noted that the background area we blend into the foreground video may not be the actual background and might contain unrelated motions. Those motions are necessary for robust motion-pattern learning. If all the background is filled with static pixels, the model will be collapsed to learn whether the region contains dynamic pixels as a shortcut. 

Besides, we have tried three variants to obtain the foreground mask $\widetilde{M}$. Though we admit the quality of our generated mask cannot be comparable to the (semi-)supervised foreground discovery methods~\cite{Croitoru_2017_ICCV,Xie_2019_CVPR}, we find all variants of FAME consistently enhance the representation ability shown in Table~\ref{tab:variants} and FAME works best among them. In addition, we test the real-time performance of FAME (\textbf{480} fps) on a single 8G NVIDIA T4 GPU at 16$\times$224$\times$224 pixels, which is negligible in terms of whole pretraining. 

\section{Experiments}
In this section, we first introduce datasets and the implementation details for our experiments. Then, we conduct a set of ablation studies to analyze and validate our FAME method quantitatively. As the evaluation, we report our results on downstream tasks: action recognition and video retrieval. 
Finally, we investigate and make sense of what the model learns with FAME qualitatively.

\subsection{Datasets}
We evaluate our method on four standard video benchmarks. Kinetics-400~\cite{carreira2017quo} is a large-scale and high-quality dataset for action recognition, which consists of around 240K video clips with 400 human action classes. We use the training set of Kinetics-400 to pretrain our model in a self-supervised manner. 
UCF101~\cite{soomro2012ucf101} and HMDB51~\cite{kuehne2011hmdb} are two smaller human action datasets, where the former contains over 13k clips covering 101 action classes and the latter annotates around 7,000 manually annotated clips with 51 action categories. Following previous methods~\cite{xu2019self, wang2020self, luo2020video, pan2021videomoco}, we use split 1 of UCF101 and HMDB51 in our experiments for downstream tasks. Also, we adopt UCF101 split 1 to conduct pre-training of our model for ablation experiments. Finally, we consider a more challenging dataset Diving48~\cite{li2018resound} for evaluation, which involves around 18k trimmed video clips of 48 dive categories. It is noted that the different diving sequences in Diving48 often occur in a similar background and primarily differ in the fine-grained motion pattern.   

\subsection{Implementation Details}
\noindent{\textbf{Self-supervised Pretraining.}}
In the stage of self-supervised training, we adopt the MoCo framework~\cite{he2020momentum, chen2020improved} as the representative of vanilla contrastive methods and apply our FAME method on MoCo framework. We select two common backbone choices, R(2+1)D-18~\cite{tran2018closer} and I3D-22~\cite{carreira2017quo}, as the 3D encoder. First, we randomly sample two different temporal clips in the same video as positive pair. Each clip consists of 16 frames with a temporal stride of 2. We spatially crop a random portion of clips and resize it to the size of $224 \times 224$ or $112 \times 112$. Then, we employ FAME to distract one out of the positive pairs. Notice that the background videos are from the clips in the same mini-batch. Next, following the prior work~\cite{feichtenhofer2021large}, we perform the basic augmentation containing random grayscale, color jittering, random horizontal flip, and random Gaussian blur. All these augmentations are temporally consistent according to \cite{qian2021spatiotemporal}. 
We pretrain the model for 200 epochs with a batch size of 64 on 8 Tesla V100 GPUs during the training phase. The SGD optimizer is adopted with the initial learning rate of $10^{-2}$ and weight decay of $10^{-4}$. We show more implementation details in the Supplementary Material.


\noindent{\textbf{Action Recognition.}} After pretraining, we initialize the backbone with the pretrained parameters except for the last fully connected layer. There are two popular protocols of action recognition to validate the self-supervised representations. One is \textit{linear probe}. The encoder is frozen, and we only train the last fully connected layer. The second one is \textit{finetune}, where we train the whole network in a supervised fashion. 
During the inference phase, we take the standard evaluation protocol~\cite{xu2019self, wang2020self, pan2021videomoco}. We uniformly sample ten $16$-frame video clips with a temporal stride of 2 from each testing video, then crop and resize them to $224 \times 224$ or $112 \times 112$. We average the prediction of each testing video clip and report Top-1 accuracy to measure the performance. 

\noindent{\textbf{Video Retrieval.}}
Without further training, we directly leverage the representation from the pretrained encoder for evaluation. Following~\cite{xu2019self, luo2020video}, we take video clips in the test set to query $k$ nearest neighbors in the training set. Specifically, we average ten uniformly sampled clips to obtain the global representation. If the category of the testing clip appears in the $k$ nearest neighbors, it counts as a hit. We report Top-$k$ recall R@k for evaluation.

\subsection{Ablation Study}
To analyze how our FAME improves self-supervised video representation learning, we conduct the following ablation studies. We choose split 1 of UCF101 as the pretrain dataset and I3D as the backbone for computational efficiency. All of the Top-1 accuracy in our ablation study is measured under the protocol of finetune.

\noindent{\textbf{Area Ratio of Foreground Region.}}
To inspect how the area ratio of the foreground region contributes to the representation quality, we ablate $\beta$ (i.e., the portion of the foreground) in the range of $\{1, 0.7, 0.5, 0.3\}$. We report the performance comparison in Table~\ref{tab:single}. Note that $\beta=1$ reverts to the baseline method without applying FAME.    
It can be observed that the results of $\beta=0.3$ and $0.5$ vastly outperform baseline by $\sim 6\%$ on both UCF101 and HMDB51. The improvement of $\beta=0.7$ is also considerable, though slightly inferior to the smaller value of $\beta$ due to insufficient background replacement. It validates our idea that replacing the static area guides the model to distill motion-aware representations and thus enhance the downstream performance. 

\begin{table}
    \centering
    \small
    \begin{tabular}{lllll}
         \bottomrule & \multicolumn{2}{c} {UCF101} & \multicolumn{2}{c} {HMDB51} \\
         $\beta$ & single & both & single & both \\\hline
         $1.0$(baseline)& \multicolumn{2}{c} {75.8} & \multicolumn{2}{c} {45.5}\\
         $0.7$ & 80.3 & 79.6 & 49.6 & 50.8 \\
         $0.5$ & 81.2 & 81.2 & 52.6 & 51.4 \\
         $0.3$ & 82.0 & 81.1 & 51.6 & 53.1 \\
        \bottomrule
    \end{tabular}
    \caption{Top-1 accuracy with $\beta$ on UCF101 and HMDB51. We denote the operating FAME on single branch (default setting) as \textit{single} and the operating FAME on both branches as
    \textit{both}.}
    \label{tab:single}
    \vspace{-0.5em}
\end{table}
\begin{table}
    \centering
    \small
    \begin{tabular}{lll}
         \bottomrule Method & Pretrain Dataset & Diving48 \\\hline
         Random Init. & \XSolidBrush & 57.4 \\
         BE~\cite{wang2021removing} & UCF101 & 58.8 \\
         \textbf{FAME(ours)} & UCF101 & 67.8 \\
         BE~\cite{wang2021removing} & Kinectics-400 & 62.4 \\
         \textbf{FAME(ours)} & Kinectics-400 & 72.9 \\
        \bottomrule
    \end{tabular}
    \caption{Top-1 accuracy on Diving48 according to updated labels (\textbf{V2}). Both methods use I3D and 16$\times$224$\times$224 pixels.}
    \label{tab:diving}
    \vspace{-0.75em}
\end{table}
\begin{table}
    \centering
    \small
    \begin{tabular}{lll}
         \bottomrule Background  & UCF101 & HMDB51 \\\hline
         none & 75.8 & 45.5 \\
         intra-video & 77.4(1.6$\uparrow$) & 47.6(2.1$\uparrow$) \\
         inter-video & 81.2(5.4$\uparrow$) & 52.6(7.1$\uparrow$) \\
        \bottomrule
    \end{tabular}
    \caption{Top-1 accuracy on UCF101 and HMDB51 in terms of intra-/inter-video background.} 
    \label{tab:intra-inter}
    \vspace{-0.5em}
\end{table}
\begin{table}
    \centering
    \small
    \begin{tabular}{ccc}
    \bottomrule
      Method  &  UCF101  &  HMDB51  \\\hline
      baseline & 75.8 & 45.5 \\
      Gauss   & 77.9 & 46.4 \\
      Seed  & 80.4 & 51.3 \\
      Grid  & 81.5 & 51.5 \\
      FAME   & 81.2 & 52.6 \\\hdashline
      Grid$\dagger$ &86.5 &58.7 \\
      FAME$\dagger$ & 88.6 & 61.1 \\
    \bottomrule
    \end{tabular}
    \caption{Top-1 accuracy of various foreground-background separation methods on UCF101 and HMDB51. $\dagger$ indicates the pretrain dataset is Kinetics-400. FAME performs best.}
    \label{tab:variants}
    \vspace{-0.5em}
\end{table}

\noindent{\textbf{Stronger background debiasing.}}
To explore whether FAME is sufficiently strong to reduce the background bias in contrastive learning, we design a stronger contrastive objective. That is, we apply FAME on both branches of MoCo and neither of the two processed video clips contains initial background information. We report the results in Table~\ref{tab:single}.
The neglectable difference in performance between both settings proves that our default setting is strong enough to learn the scene-debiased representations. 

\noindent{\textbf{Background source.}}
Besides the foreground ratio, we also investigate how the source of background affects the representation ability to capture the motion. 
Specifically, we aim to explore whether the performance would change dramatically using the background in the same video instead of other videos. We perform an experiment where we merge the foreground of one video with the background sampled at different timestamps of the video itself. As shown in Table~\ref{tab:intra-inter}, we find that using the background from intra-video slightly boosts the baseline with 1.6\% and 2.1\% improvement on UCF101 and HMDB51 and the introduction of other videos' backgrounds brings further improvement, i.e., 5.4\% and 7.1\% gain on UCF101 and HMDB51. In general, the intra-video background is almost the same as the original one, while the inter-video background is quite distinct. Thus, it is consistent with our intuition that the modification from the intra-video is not adequate to mitigate background bias while replacing the background with diverse scenes better strengthens motion pattern learning.

\begin{table*}[!h]
\centering
\small
    \begin{tabular}{llllllll}
        \toprule
        Method & Backbone & Pretrain Dataset & Frames & Res. & Freeze & UCF101 & HMDB51 \\\hline
        CBT~\cite{sun2019learning} & S3D & Kinetics-600 & 16 & 112 & \Checkmark & 54.0 & 29.5 \\
        CCL~\cite{kong2020cycle} & R3D-18 & Kinetics-400 & 16 & 112  & \Checkmark & 52.1 & 27.8 \\
        MemDPC~\cite{han2020memory} & R3D-34 & Kinetics-400 & 40 & 224 & \Checkmark & 54.1 & 30.5\\
        RSPNet~\cite{chen2021rspnet} & R3D-18 & Kinetics-400 & 16 & 112  & \Checkmark & 61.8 & \textbf{42.8} \\
        MLRep~\cite{qian2021enhancing} & R3D-18 & Kinetics-400 & 16 & 112 & \Checkmark & 63.2 & 33.4 \\ 
        \textbf{FAME (Ours)} & R(2+1)D &  Kinetics-400 & 16 & 112 & \Checkmark & \textbf{72.2} &	42.2 \\
        \hline
        \hline
        VCP~\cite{luo2020video} & R(2+1)D & UCF101 & 16 & 112  & \XSolidBrush & 66.3 & 32.2 \\
       
        PRP~\cite{yao2020video} & R(2+1)D & UCF101 & 16 & 112 & \XSolidBrush & 72.1 & 35.0\\
         TempTrans~\cite{jenni2020video} & R(2+1)D & UCF101 & 16 & 112  & \XSolidBrush & 81.6 & 46.4 \\
         3DRotNet~\cite{jing2018self} & R3D-18 & Kinetics-400 & 16 & 112  & \XSolidBrush &62.9 &33.7\\
         Spatio-Temp~\cite{wang2019self} & C3D & Kinetics-400 & 16 & 112 & \XSolidBrush & 61.2 & 33.4
        \\ 
   
        Pace Prediction~\cite{wang2020self} & R(2+1)D & Kinetics-400 & 16 & 112  & \XSolidBrush & 77.1 & 36.6
        \\
        SpeedNet~\cite{benaim2020speednet} & S3D-G & Kinetics-400 & 64 & 224  & \XSolidBrush & 81.1 & 48.8\\
        VideoMoCo~\cite{pan2021videomoco} & R(2+1)D & Kinetics-400 & 32 & 112 & \XSolidBrush & 78.7 & 49.2
        \\
        RSPNet~\cite{chen2021rspnet} & R(2+1)D & Kinetics-400 & 16 & 112  & \XSolidBrush & 81.1 &44.6 \\

        MLRep~\cite{qian2021enhancing} & R3D-18 & Kinetics-400 & 16 & 112 & \XSolidBrush & 79.1 & 47.6 \\ 
        
        ASCNet~\cite{huang2021ascnet} & R3D-18 & Kinetics-400 & 16 & 112 & \XSolidBrush & 80.5 & 52.3 \\
        SRTC~\cite{zhang2021incomplete} & R(2+1)D &  Kinetics-400 & 16 & 112 & \XSolidBrush & 82.0 & 51.2 \\
        \textbf{FAME (ours)} & R(2+1)D & Kinetics-400 & 16 & 112 & \XSolidBrush &	 \textbf{84.8}	& \textbf{53.5} \\
        \hline
        DSM~\cite{wang2021enhancing}& I3D & Kinetics-400 & 16 & 224 & \XSolidBrush & 74.8 & 52.5 \\ 

        BE~\cite{wang2021removing}& I3D & Kinetics-400 & 16 & 224 & \XSolidBrush & 86.8 & 55.4 \\ 
        
        \textbf{FAME (ours)} & I3D & Kinetics-400 & 16 & 224 & \XSolidBrush & \textbf{88.6} & \textbf{61.1} \\
        \bottomrule
    \end{tabular}
    \caption{Comparison with the existing self-supervised video representation learning methods for action recognition on UCF101 and HMDB51. To compare fairly, we list each work's setting, including backbone architecture used, pretrain dataset and spatial-temporal resolution. Freeze (tick) indicates linear probe, and no freeze (cross) means finetune.}
    \vspace{-0.5em}
    \label{tab:recognition}
\end{table*}

\begin{table*}
\centering
\small
    \begin{tabular}{lllllll}
        \toprule
        \multirow{2}{*}{Method} & \multirow{2}{*}{Backbone}  & \multicolumn{5}{c}{R@k}  \\\cline{3-7} 
        &&R@1 & R@5 & R@10 & R@20 & R@50\\\hline
        SpeedNet~\cite{benaim2020speednet} & S3D-G &  13.0 & 28.1 & 37.5 & 49.5 & 65.0\\
        TempTrans~\cite{jenni2020video} & R3D-18 & 26.1 & 48.5 & 59.1 & 69.6 & 82.8 \\
        MLRep~\cite{qian2021enhancing} & R3D-18 &  41.5 & 60.0 & 71.2 & 80.1  & - \\
        GDT~\cite{patrick2020multi} & R(2+1)D & 57.4 & 73.4 & 80.8 & 88.1 & 92.9 \\
        ASCNet~\cite{huang2021ascnet} & R3D-18 &  58.9 & 76.3 & 82.2 & 87.5 & 93.4\\
        \textbf{FAME (ours)} & R(2+1)D & 64.6 & 77.7 & 82.9 & 87.6 & 94.2\\
        \bottomrule
    \end{tabular}
    \caption{Comparison with the existing self-supervised video representation learning methods for video retrieval. All methods are pretrained on Kinetics-400. We report the Top-$k$ recall R@k when k=1, 5, 10, 20, 50 on UCF101.}
    \label{tab:retrieval}
    \vspace{-0.75em}
\end{table*}
\begin{figure}[!h]
    \centering
    \includegraphics[width=\linewidth]{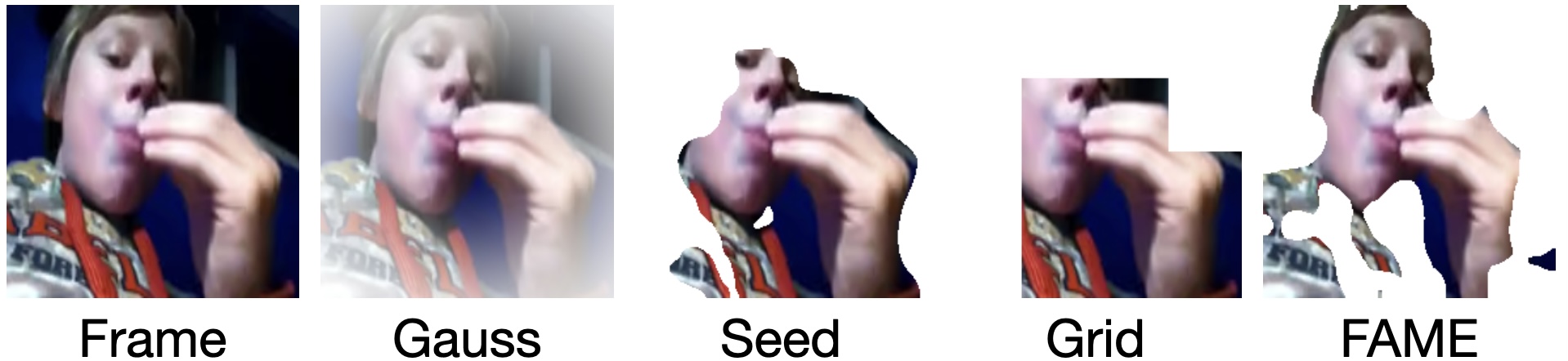}
    \caption{The illustration about FAME and three variants.}
    \label{fig:variants}
    \vspace{-0.75em}
\end{figure}
\noindent{\textbf{Variants of Foreground Mask.}}
To verify that emphasizing moving foreground advances the motion understanding in contrastive framework, we devise three variants of foreground mask: (i)  Gauss: we adopt a 2D Gaussian kernel matrix as the foreground mask. It derives from the assumption that videos are shot in the object-centric form. 
(ii) Seed: we just take the seed region $S$ to characterize the foreground. 
(iii) Grid: the video is split into $4 \times 4$ grids spatially. We count the sum of $S$ in each grid and take the greatest eight grids as the foreground area. 
A brief illustration is displayed in Fig.~\ref{fig:variants}. 
We compare FAME with these three variants in Table~\ref{tab:variants}. First, we note that all variants improve the baseline by a large margin, demonstrating the benefit of the introduction of different backgrounds. Furthermore, refining the foreground mask from Gauss, Seed, Grid to FAME continually increases the action recognition performance. Interestingly, we notice that Grid outperforms FAME a little on UCF101. 
We conjecture that since both the pretrain dataset and downstream dataset are UCF101, similar backgrounds that occurred might be leveraged as a shortcut. To delve into this phenomenon, we carry out an extra experiment on another pretrain dataset Kinetics-400. Top-1 accuracy of Grid variant is over 2\% lower than FAME on both UCF101 and HMDB51. 
It indicates that a meticulous segmentation mask instead of a rough grid box is more effective in facilitating generalization ability when transferring the motion-aware representations to different downstream benchmarks.

\subsection{Evaluation on Downstream Tasks}

\noindent{\textbf{Action Recognition on UCF101 and HMDB51.}}
To verify the effectiveness of the proposed method, we compare our method with the prior arts. In Table~\ref{tab:recognition}, we report Top-1 accuracy on UCF101 and HMDB51. For a fair comparison, we do not report methods with a deeper backbone or non-single modality, e.g., optical flow, audio, and text.

Our method obtains the best result on UCF101 and a comparable result on HMDB51 in the linear probe setting. FAME beats MLRep~\cite{qian2021enhancing} by a large margin, i.e., about 9.0\% gain on both UCF101 and HMDB51, where MLRep carefully designs the multi-level feature optimization and temporal modeling. The outstanding performance verifies that our method captures the moving foreground patterns without further finetune. 

In the finetune protocol, FAME with R(2+1)D backbone achieves the best result on UCF101 and HMDB51. It shows that FAME learns the scene-debiased and motion-aware representations on the Kinetics-400 dataset, which would generalize better to downstream datasets. Remarkably, our simple formulation outperforms SRTC~\cite{zhang2021incomplete} by $2.8\%$ and $2.3\%$ with the same backbone R(2+1)D, despite its two additional sub-loss terms regularizing the self-supervised pretraining. Notably, we share similar motivation with BE~\cite{wang2021removing}, which directly adds a static frame to every other frame and regards this distracting video as the positive pair to the original video. Such subtle disturbance cannot sufficiently mitigate the static background bias, confirmed by the experimental results. When using the same backbone I3D, our FAME outperforms BE by 1.8\% and 5.7\% on UCF101 and HMDB51, respectively. It proves that our method can better highlight the motion patterns.

\noindent{\textbf{Video Retrieval on UCF101 and HMDB51.}}
We report the performance comparison on the video retrieval task in Table~\ref{tab:retrieval}. Our method achieves significant performance gain from R@1 to R@50. Remarkably, though ASCNet~\cite{huang2021ascnet} devises two particular tasks to learn the appearance and speed consistency, we still gain 6.7\% improvement on Top-1 retrieval accuracy only through fabricating motion-aware positive pairs, which demonstrates our methods can recognize the action semantics more precisely.

\noindent{\textbf{Evaluation on Diving48.}}
Besides common action recognition benchmarks, we finetune and test our FAME on a more challenging fine-grained dataset Diving48 and report the results in Table~\ref{tab:diving}. 
In Diving48, since the static backgrounds are not strongly related to the fine-grained diving label, our motion-aware representations can strongly benefit the action recognition. FAME can boost a randomly initialized model by 15.5\% when pretraining on Kinetics-400. 
In comparison, no matter whether the pretrain dataset is UCF101 or Kinetics-400, BE is far less effective than FAME. This is because BE does not construct the motion-aware positive pairs where background features are still more similar in contrast to the motion counterparts. The result on Diving48 indicates FAME can indeed make the model perceive the long-term motion patterns and hinder the scene bias.

\subsection{Visualization Analysis}

To better demonstrate the effectiveness of FAME, we visualize the CAAM~\cite{baek2020psynet} in Fig.~\ref{fig:caam}. By comparison, the model learned by FAME enhances the activation on the moving foreground area and suppresses the background area. For example, in the second column of Fig.~\ref{fig:caam}, FAME precisely captures two baseball players in the court, while the vanilla contrastive method displays a dispersed highlight map and fails to attend to the motion area. 

Moreover, compared to vanilla contrastive method, the distribution of temporal kernel learned by FAME is more scattered with larger variance as shown in Fig.~\ref{fig:violin}. Surprisingly, the shape of FAME's temporal kernel weights is similar to supervised learning, showing that via FAME, contrastive learning can well grasp action semantics.
In light of the aforementioned evidence, 
we safely conclude that guided by the strong motion inductive augmentation like FAME, contrastive learning can also prevent background cheating and pay attention to the motion patterns.   

\section{Conclusion}
In this work, we propose a new Foreground-background Merging (FAME) method to alleviate the background bias in self-supervised video representation learning.
Via Foreground-background Merging, we augment the original video by fusing the original foreground with other videos' backgrounds. When forcing the backbone model to learn semantically consistent representation between the original video and the fused video, the model can learn the scene-debiased and motion-aware representations of videos. Experimental results on a bunch of downstream tasks manifest the effectiveness of our method. 

While our work shows some promising results, there are still some limitations. One is that the quality of foreground extraction is not stable, especially when foreground and background have no significant differences in the color distribution or the camera is dynamically moving. Besides, the foreground area ratio is now fixed by hyper-parameter $\beta$. It would be better to set an adaptive foreground area ratio. 
\section*{Acknowledgment}
This work was supported in part by the National Natural Science Foundation of China under Grant 61932022, Grant 61720106001, Grant 61971285, Grant 61831018, Grant 61871267, Grant T2122024, and in part by the Program of Shanghai Science and Technology Innovation Project under Grant 20511100100. 
{\small
\bibliographystyle{ieee_fullname}
\bibliography{egbib}
}
\newpage
\appendix
\section{More Implementation Details}
\subsection{Self-supervised Pretraining Details.}

In the pretraining stage, we adopt the SGD optimizer with the initial learning rate of 0.01 and weight decay of $10^{-4}$, and we decay the learning rate by 0.1 at epoch 120 and 180. For the implementation of MoCo, we closely follow the parameter setting in~\cite{chen2020improved}. The number of the negative queue is set to 65536 for Kinetics-400, and 2048 for UCF101, respectively. We also swap the key/queue samples so that each sample can generate the gradient for optimization. The momentum of updating the key encoder is 0.999, and the temperature hyper-parameter $\tau$ is 0.1. We use a 2-layer MLP projection head.
\subsection{Augmentation Details.}
We perform data augmentation using Kornia package~\cite{riba2020kornia}.
In the pretraining and finetune phase, we crop 224$\times$224 or 112$\times$112 pixels from a video with RandomResizedCrop, which randomly resizes the input area between a lower bound and upper bound. We set the bound as [0.2, 1]. Then, the basic augmentation set consists of RandomGrayscale (probability 0.2), ColorJitter (probability 0.8, \{brightness, contrast, saturation, hue\} = \{0.4, 0.4, 0.4, 0.1\}), RandomHorizontalFlip (probability 0.5) and RandomGaussianBlur (probability 0.5, the kernel with radius 23 and standard deviation $\in$ [0.1, 2.0]). In the linear probe stage, we take a simpler augmentation setting instead. We only apply RandomResizedCrop with the bound [0.2, 1] and RandomHorizontalFlip (probability 0.5).

\subsection{More Details on Action Recognition.}
 In the finetune stage, the SGD optimizer is adopted with the initial learning rate of 0.025 and weight decay of $10^{-4}$. We finetune the model for 150 epochs with a batch size of 128 on 4 Tesla V100 GPUs. We decay the learning rate by 0.1 at epoch 60 and 120. Besides, we add the dropout layer before the last fully connected layer. We set dropout rate 0.7 for UCF101 and 0.5 for HMDB51, respectively. 
 
We train the last fully connected layer in the linear probe with the initial learning rate of 5 and weight decay of 0. We finetune the model for 100 epochs with a batch size of 128 on 4 Tesla V100 GPUs. We decay the learning rate by 0.1 at epoch 60 and 80. Besides, We $\mathcal{L}_2$ normalize the embeddings before the last fully connected layer. 
\section{More Visualization of FAME}
\begin{figure}
    \centering
    \includegraphics[width=\linewidth]{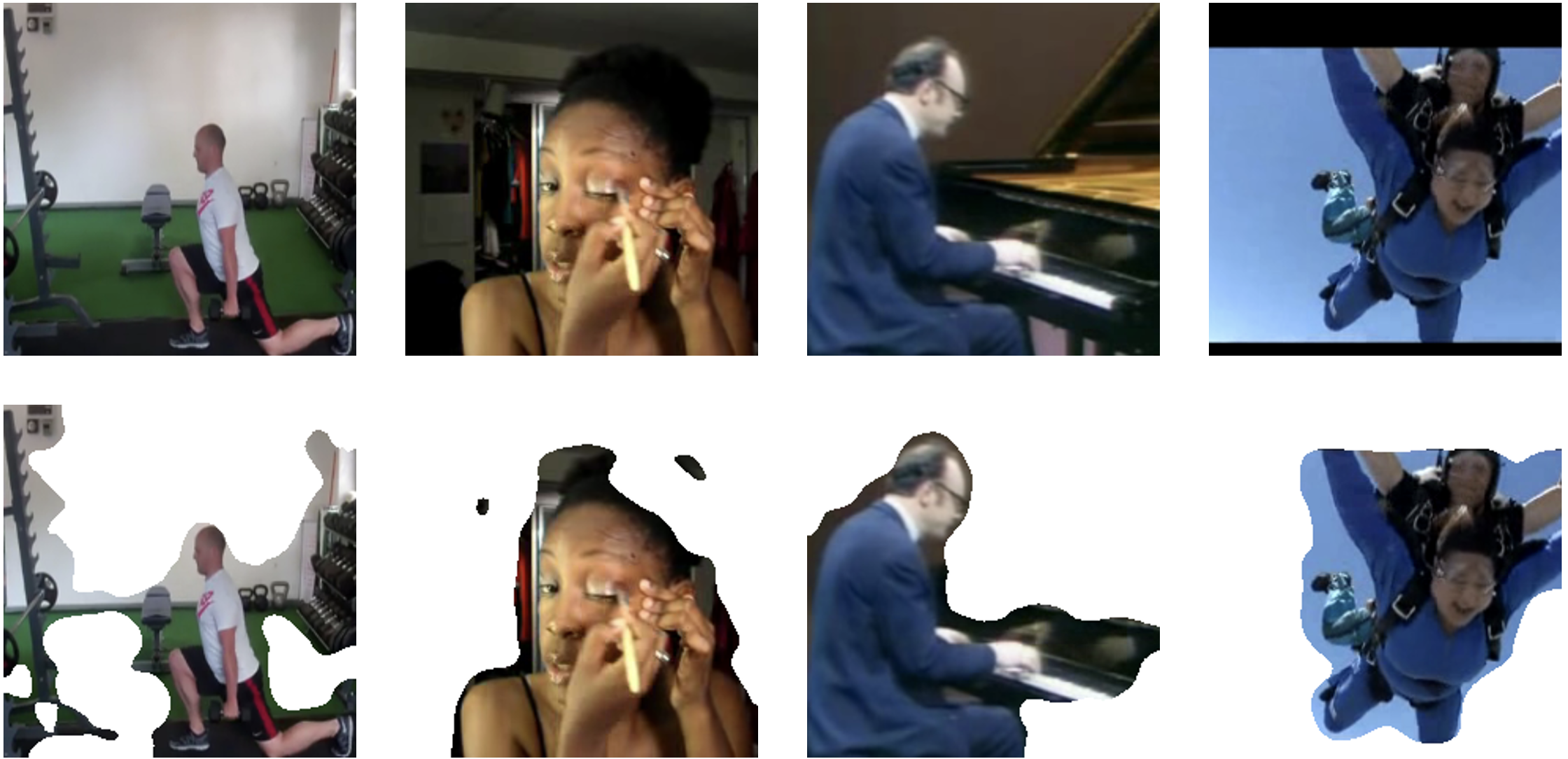}
    \caption{Illustration of FAME visualization. The first row is the video frame while the second row is the foreground mask FAME generates.}
    \label{fig:FAME}
\end{figure}
In Figure~\ref{fig:FAME}, we show more foreground masks obtained from FAME. We show that FAME can discover most regions of the foreground objects and remove the monotonous backgrounds. 

\section{CAM Visualization}

\begin{figure}
    \centering
    \includegraphics[width=\linewidth]{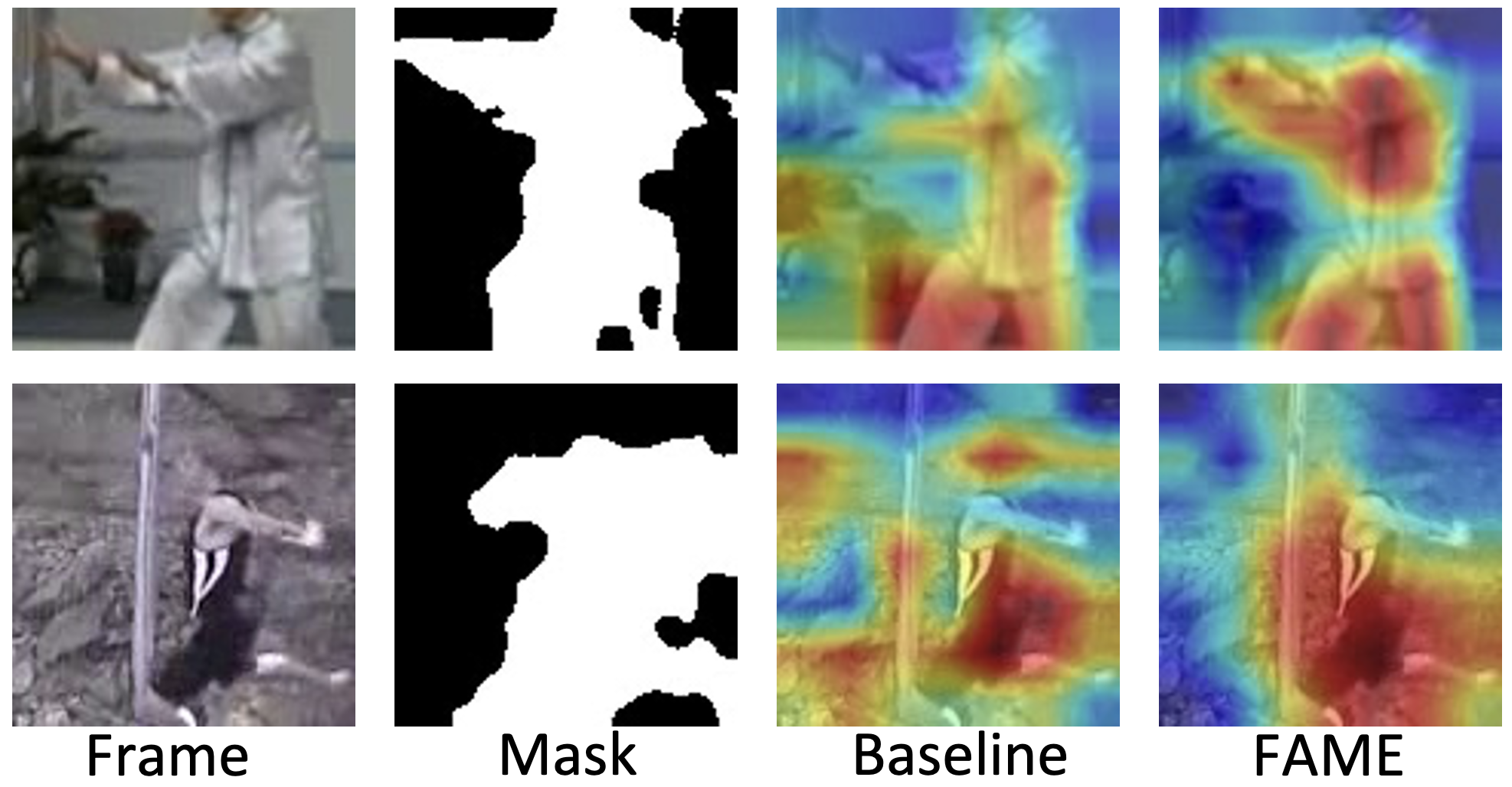}
    \caption{Class activation maps (CAM) visualization. Red areas indicate the important areas for the model to predict the action class. Comparing to baseline, FAME resists the impact of background and highlights the motion areas.}
    \label{fig:cam}
\end{figure}

Besides CAAM visualization, we provide the CAM~\cite{zhou2016learning} visualization in Figure~\ref{fig:cam}.
With that, we can spot the contribution of each area and find crucial regions for discriminating the specific action class. We find that when integrated with FAME, the model can focus on moving foreground area rather than background context. For example, in the first row of Figure~\ref{fig:cam}, FAME precisely captures the moving upper and lower body when the man is practicing TaiChi, while the baseline displays a dispersed highlight map and fails to attend to the motion area. In addition, we illustrate that the CAM activation map can almost overlap with the foreground mask generated by FAME. It testifies that our strong motion inductive augmentation guides the model to perceive the motion patterns and hinder the background bias. 

\section{Visualization of Video Retrieval}
\begin{figure}
    \centering
    \subfloat[BandMarching]{\includegraphics[width=0.25\linewidth]{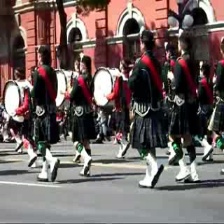}
    \hspace{2mm}
    \includegraphics[width=0.25\linewidth]{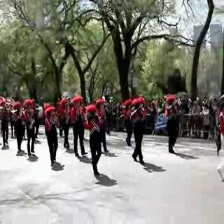}
    \includegraphics[width=0.25\linewidth]{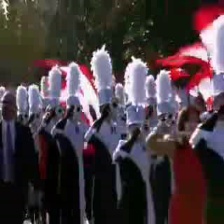}
    \includegraphics[width=0.25\linewidth]{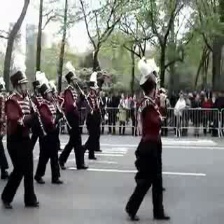}}
    \hspace{4mm}\\
    \subfloat[BenchPress]{\includegraphics[width=0.25\linewidth]{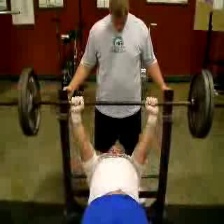}
    \hspace{2mm}
    \includegraphics[width=0.25\linewidth]{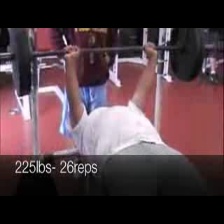}
    \includegraphics[width=0.25\linewidth]{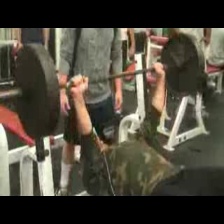}
    \includegraphics[width=0.25\linewidth]{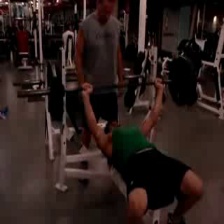}}
    \hspace{4mm}\\
    \subfloat[HorseRace]{\includegraphics[width=0.25\linewidth]{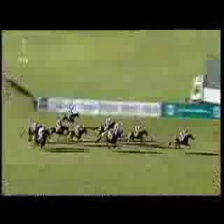}
    \hspace{2mm}
    \includegraphics[width=0.25\linewidth]{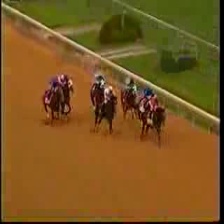}
    \includegraphics[width=0.25\linewidth]{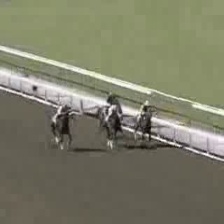}
    \includegraphics[width=0.25\linewidth]{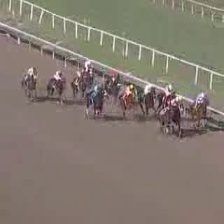}}
    \hspace{4mm}\\
    \subfloat[LongJump]{\includegraphics[width=0.25\linewidth]{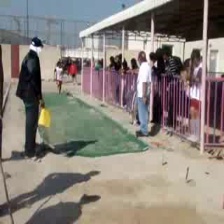}
    \hspace{2mm}
    \includegraphics[width=0.25\linewidth]{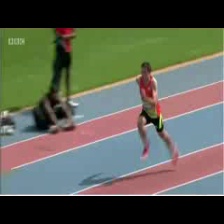}
    \includegraphics[width=0.25\linewidth]{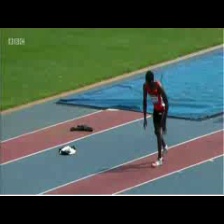}
    \includegraphics[width=0.25\linewidth]{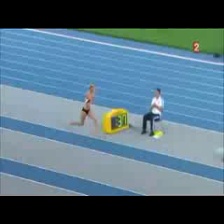}
    \label{fig:longjump}}
    \caption{Visualization results of video retrieval.The first column is the video frame of query instances. The rightmost three columns are Top-3 nearest retrieval results.}
    \label{fig:RET}
\end{figure}

In Figure~\ref{fig:RET}, we demonstrate the results of video retrieval. After pretraining the model on Kinetics-400, we conduct the video retrieval experiment on UCF101. The results show that our model can retrieve diverse video samples that share the same action semantics with the query, regardless of the background context. For example, in Fig~\ref{fig:longjump}, the query sample contains the action in the sandpit, and our model could retrieve the long jump samples in the standard stadium. Though the backgrounds in the query and retrieved videos are quite different, our model achieves accurate retrieval by attending to the dynamic motions and understanding the true action semantics.

\section{More Results on Something-something V2}
We finetune our pretained model on Something-Something V2~\cite{goyal2017something}. We obtain 53.3\% Top-1 accuracy  with R(2+1)D, which beats RSPNet~\cite{chen2021rspnet} under same resolution. 
\end{document}